\documentclass[
]{ceurart}

\usepackage{listings}
\usepackage{todonotes}
\usepackage{tikz}

\begin{document}

\copyrightyear{2022}
\copyrightclause{Copyright for this paper by its authors.
  Use permitted under Creative Commons License Attribution 4.0
  International (CC BY 4.0).}

\conference{CLEF 2022: Conference and Labs of the Evaluation Forum, 
    September 5--8, 2022, Bologna, Italy}

\newcommand{\MacquarieUniversity}{Macquarie University}
\newcommand{\MQ}{MQ}


\title{Query-focused Extractive Summarisation for Biomedical and COVID-19 Complex Question Answering}
\title[mode=alt]{\MacquarieUniversity's Participation at BioASQ10 Synergy and BioASQ10b Phase~B}

\author[1]{Diego Moll\'a}[%
orcid=0000-0003-4973-0963,
email=diego.molla-aliod@mq.edu.au,
url=https://researchers.mq.edu.au/en/persons/diego-molla-aliod,
]
\address[1]{Macquarie University, Australia}

\begin{abstract}
This paper presents \MacquarieUniversity's participation to the two most recent BioASQ Synergy Tasks (as per June 2022), and to the BioASQ10 Task~B (BioASQ10b), Phase~B. In these tasks, participating systems are expected to generate complex answers to biomedical questions, where the answers may contain more than one sentence. We apply query-focused extractive summarisation techniques. In particular, we follow a sentence classification-based approach that scores each candidate sentence associated to a question, and the $n$ highest-scoring sentences are returned as the answer. The Synergy Task corresponds to an end-to-end system that requires document selection, snippet selection, and finding the final answer, but it has very limited training data. For the Synergy task, we selected the candidate sentences following two phases: document retrieval and snippet retrieval, and the final answer was found by using a DistilBERT/ALBERT classifier that had been trained on the training data of BioASQ9b. Document retrieval was achieved as a standard search over the CORD-19 data using the search API provided by the BioASQ organisers, and snippet retrieval was achieved by re-ranking the sentences of the top retrieved documents, using the cosine similarity of the question and candidate sentence. We observed that vectors represented via sBERT have an edge over tf.idf. BioASQ10b Phase B focuses on finding the specific answers to biomedical questions. For this task, we followed a data-centric approach. We hypothesised that the training data of the first BioASQ years might be biased and we experimented with different subsets of the training data. We observed an improvement of results when the system was trained on the second half of the BioASQ10b training data.
\end{abstract}


\maketitle

\section{Introduction}

The BioASQ challenge\footnote{\url{http://www.bioasq.org/}} organises shared tasks on biomedical semantic indexing and question answering. In this paper, we present \MacquarieUniversity's participation in several of these tasks.\footnote{Code related to this paper is available at \url{https://github.com/dmollaaliod/bioasq10b-public} and \url{https://github.com/dmollaaliod/bioasq10-synergy-public}.} 

The Synergy tasks aim to evaluate technologies useful for the development of an end-to-end question answering (QA) system for questions about COVID-19 asked by biomedical experts. In particular, the Synergy tasks evaluate the quality of document retrieval over a snapshot of CORD-19 \cite{wang-etal-2020-cord}, snippet retrieval, and the generation of ``ideal answers'' that may contain multiple sentences. We present our participation in the second BioASQ9 Synergy task that ran between May and June 2021, and the BioASQ10 Synergy task that ran between December 2021 and February 2022.

Task B of BioASQ focuses on biomedical semantic QA. Similar to the Synergy tasks, several technologies corresponding to components of an end-to-end QA system are evaluated. In contrast with the Synergy tasks, Task B of BioASQ has two distinct phases. Phase A evaluates the quality of document and snippet retrieval on a snapshot of PubMed\footnote{\url{https://pubmed.ncbi.nlm.nih.gov/}}, whereas Phase B, given a question, its question type (``summary'', ``factoid'', ``yesno'', ``list'') , and a list of candidate snippets, evaluates the system's ability to find short answers (``exact answers'') and long, possibly multi-sentence answers (``ideal answers''). Figure~\ref{fig:Examples} shows an example of a question and its question type, a correct snippet for the question, a correct exact answer, and a correct ideal answer. We present our participation in Task B, Phase B of BioASQ10, that ran between March and May 2022 (henceforth BioASQ10b, Phase B). 

\begin{figure}
    \centering
    \begin{description}
        \item[Question] Orteronel was developed for treatment of which cancer?
        \item[Type] factoid
        \item[Snippet] Pooled-analysis was also performed, to assess the effectiveness of agents targeting the androgen axis via identical mechanisms of action (abiraterone acetate, orteronel).
        \item[Exact answer] castration-resistant prostate cancer
        \item[Ideal answer] Orteronel was developed for treatment of castration-resistant prostate cancer.
    \end{description}
    \caption{An example question with its question type, a relevant snippet, an exact answer, and a correct ideal answer, extracted from the training data of BioASQ10b}
    \label{fig:Examples}
\end{figure}

All of our contributions to the above tasks are based on a common question-answering architecture that we will describe in Section~\ref{sec:architecture}. Section~\ref{sec:synergy} presents our participation in the Synergy tasks. Section~\ref{sec:bioasq10b} presents our participation in BioASQ10b, Phase B. Finally, Section~\ref{sec:conclusions} concludes this paper.

\section{Question Answering Architecture}\label{sec:architecture}

\begin{figure}
    \centering
    \begin{tikzpicture}[scale=0.38]
    \footnotesize
% input
    \filldraw[fill=blue!20!white, draw=blue!40!black] (0,0) rectangle (1,5) (0,1) -- (1,1) (0,3) -- (1,3) (0,4) -- (1,4);
    \filldraw[fill=red!20!white, draw=red!40!black] (0,-7) rectangle (1,-2) (0,-6) -- (1,-6) (0,-4) -- (1,-4) (0,-3) -- (1,-3);
    \draw (-1,2.5) node[rotate=90] {sentence};
    \draw (-1,-4.5) node[rotate=90] {question};

% word embeddings
    \draw (4,-1) node [circle,draw,align=center,text width=1cm] (em) {BERT};
    \draw (8.5,6) node {word embeddings};
    \filldraw[fill=blue!20!white, draw=blue!40!black] (7,0) rectangle (10,5) (7,1) -- (10,1) (7,3) -- (10,3) (7,4) -- (10,4) (8,0) -- (8,5) (9,0) -- (9,5);

    \draw[->] (1,2.5) -- (em);
    \draw[->] (1,-4.5) -- (em);

    \draw[->] (em) -- (7,2.5);
    \draw (14,2.5) node [circle,draw,align=center,text width=1cm] (sr) {Mean};
    \draw (18,6) node {sentence embeddings};
    \filldraw[fill=blue!20!white, draw=blue!40!black] (18,0) rectangle (19,5) (18,1) -- (19,1) (18,3) -- (19,3) (18,4) -- (19,4);

    \draw[->] (10,2.5) -- (sr);
    \draw[->] (sr) -- (18,2.5);

    \draw[->] (19,2.5) -- (25,2.5);

% hidden layer
    \filldraw[fill=blue!20!white, draw=blue!40!black] (25,0) rectangle (26,5);
    \draw (28,2.5) circle[radius=1] (27.5,2) -- (28,2) -- (28.5,3);
    \draw (28,4) node {relu};

    \filldraw[fill=black!20!white, draw=black] (30,0) rectangle (31,5);

    \draw[->] (26.2,2.5) -- (27,2.5);
    \draw[->] (29,2.5) -- (30,2.5);

% final layer
     \draw (33,2.5) node[circle,draw] (sig) {$\int$};
\filldraw[fill=black!20!white, draw=black] (35,2) rectangle (36,3);
    \draw (33,4) node[text width=1cm] {sigmoid};

    \draw[->] (31,2.5) -- (sig);
    \draw[->] (sig) -- (35,2.5);
    
% position
    \filldraw[fill=green!20!white, draw=green!40!black] (0,6) rectangle (1,7);
    \draw (1,6.5) -- (24,6.5);
    \draw[->] (24,6.5) |- (25,5.5);
    \filldraw[fill=green!20!white, draw=green!40!black] (25,6) rectangle (26,5);
    \draw (1,7.7) node {sentence position};
  \end{tikzpicture}

    \caption{Architecture of the question answering system used for BioASQ9b, Phase B.}
    \label{fig:bioasq9b}
\end{figure}
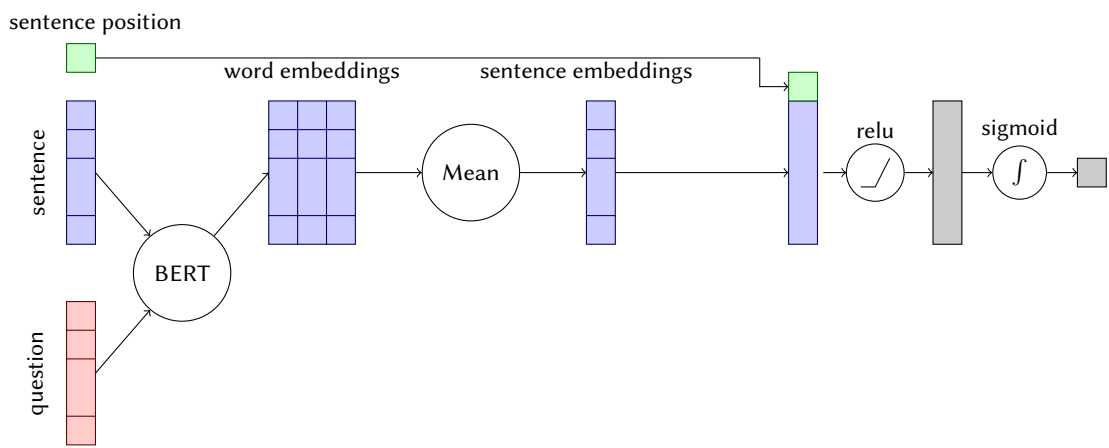

The question-answering system that is the focus of our participation in all of the tasks presented in this paper is based on query-focused extractive summarisation. The architecture of the system is illustrated in Figure~\ref{fig:bioasq9b}, and follows the classification set up proposed by~\cite{mollabioasq9b}.

The query-focused summarisation system takes the question, a candidate sentence, and the sentence position\footnote{The sentence position was incorporated as an absolute number: 1, 2, \ldots $n$, where $n$ is the total number of input sentences. We chose to include the sentence position as earlier experiments in past BioASQ years showed an improvement of the results.}, and calculates a sentence score. The system computes the word embeddings of the question and candidate sentence using a BERT architecture \cite{devlin-etal-2019-bert}. In particular, for the BioASQ9 Synergy task 2 we used ALBERT \cite{albert}, which was the best-performing system in \cite{mollabioasq8b}'s participation in BioASB8b\footnote{At the time of training the system for the BioASQ9 Synergy task 2, the final results of BioASQ9 had not been released yet.}. For the BioASQ10 Synergy task, we used DistilBERT \cite{distilbert}, which performed very well in \cite{mollabioasq9b}'s participation in BioASQ9b, and even outperformed BioBERT \cite{Lee:2019}. For BioASQ10, Phase B, we also used DistilBERT. Average pooling is then used to merge the word embeddings of the candidate sentence into the sentence embeddings. The sentence position is then concatenated to the sentence embeddings, and an additional intermediate dense layer is added. A final classification layer predicts the sentence score.

The question and the sentence were fed to BERT in the same way as defined by the creators of BERT \cite{devlin-etal-2019-bert}. That is, the input consisted of an initial ``[CLS]'' token, followed by the question text, then a ``[SEP]'' token that indicates a new sentence, and finally the candidate sentence text. This information was passed to BERT, indicating the question and the candidate sentence as two separate text segments.

The classification labels used for training the system were automatically generated from the training data, based on the ROUGE score of the candidate sentence with respect to the annotated ideal answer. In particular, given a particular question, the top 5 sentences according to their ROUGE score were labelled as 1, and the rest were labelled as 0. For the Synergy tasks we used the BioASQ9b training data, whereas for BioASQ10b, Phase B, we used the BioASQ10b training data.

We used the pre-trained ALBERT and DistilBERT models available by Huggingface\footnote{\url{https://huggingface.co/} --- For ALBERT, we used `albert-xxlarge-v2'. For DistilBERT, we used `distilbert-base-uncased'.}. These models were frozen during training, so that only the weights of the additional layers shown in Figure~\ref{fig:bioasq9b} were updated.

\section{The Synergy Tasks}\label{sec:synergy}

This section describes the systems that participated in the Synergy task 2 of BioASQ9, and the Synergy task of BioASQ10 (in this paper, we will use the collective expression ``the Synergy tasks'' to refer to these). The Synergy task 2 of BioASQ9 ran in 2021 but the results were not made available at the time of the paper submission deadline for BioASQ9. For this reason, we are describing the system in this paper.

Our participation in the Synergy tasks share the same question answering system architecture described in Section~\ref{sec:architecture}. The only difference between the two Synergy tasks is, as mentioned in Section~\ref{sec:architecture}, that the BioASQ9 Synergy 2 system used ALBERT, whereas the BioASQ10 Synergy system used DistilBERT. In both cases, the system was trained with the training data of BioASQ9b.

To generate the candidate sentences required by the question answering system, we followed this procedure:

\begin{enumerate}
    \item Retrieve the most relevant documents as described in Section~\ref{sec:docs};
    \item Split the retrieved documents into sentences and select the candidate sentences as described in Section~\ref{sec:snippets}.
\end{enumerate}

\subsection{Document Retrieval}
\label{sec:docs}

The relevant documents were retrieved using the search API provided by the organisers of the BioASQ Synergy task. This API is based on a Web service that accepts a query and returns a JSON data structure. We simply used the unmodified question as the search query. In subsequent work we are exploring pre-processing and fine-tuning steps to improve the quality of the Document Retrieval stage.

The final runs submitted consist of the top 10 documents, after removing those that were in previous feedback, to conform with the submission requirements.

\subsection{Snippet Retrieval}
\label{sec:snippets}

Every sentence from every retrieved document was a candidate snippet. This includes sentences from documents that were retrieved but were not submitted in the Document Retrieval runs.
We then experimented with the combination of 2 dimensions to re-rank the candidate snippets, for a total of 4 different approaches.

\vspace{1ex}

The first dimension was based on the calculation of the similarity between the question and candidate snippet. We experimented with the following two options:

\paragraph{TfidfCosine.}
We represented the question and candidate sentences using tf.idf. Each candidate sentence was then scored based on the cosine similarity between the question vector and the sentence vector.

\paragraph{sBERTCosine.}
We used sBERT \cite{reimers-2019-sentence-bert} to represent the question and the candidate sentences, and to determine the similarities between the question and the sentences. We used the default set up for sBERT, which computes the cosine similarity between the question vector and the sentence vector.

\vspace{1ex}

The second dimension was based on the criteria used for the final ranking of the candidate sentences. We experimented with local sorting and global sorting.

\paragraph{LocalSorting.}
For every relevant document, we extracted the top 3 sentences according to the cosine similarity approaches described above. The final list of sentences was composed of the top 3 sentences from the top document, followed by the top 3 sentences of the second document, and so on.

\paragraph{GlobalSorting.}
In contrast to the local sorting approach, all sentences of all documents were now sorted according to their cosine similarity with the question, regardless of what document the snippets were obtained from.

\vspace{1ex}

The final runs submitted consist of the first 10 snippets, after removing those that were in previous feedback, to conform with the submission requirements.

\subsection{Answer Generation}\label{sec:qa}

As mentioned above, the question, candidate sentences, and sentence position were fed to the system illustrated in Figure~\ref{fig:bioasq9b}. The sentence position was simply the unnormalised position of the sentence within the list of snippets, after the snippets have been ranked as described in Section~\ref{sec:snippets}. Given a question, the top-scoring $n$ sentences according to the scores produced by the QA system were combined to form the final answer. These sentences were presented in the order of appearance in the list of snippets. The value of $n$ was based on the question type and is shown in Table~\ref{tab:valueofn}.
\begin{table}
    \centering
    \caption{Number of sentences selected, for each question type}
    \label{tab:valueofn}
    \begin{tabular}{ccccc}
    & \textbf{Summary} & \textbf{Factoid} & \textbf{Yesno} & \textbf{List} \\
    \midrule
    \textbf{n}     &  6 & 2 & 2 & 3\\
    \end{tabular}
\end{table}

\subsection{Results of the Synergy Tasks}\label{sec:synergyruns}

This section describes the results of the runs submitted to the Synergy tasks.

Table~\ref{tab:synergydocs} shows the F1 score of the documents returned by our systems. As mentioned in Section~\ref{sec:docs}, these documents were found by submitting the unmodified question as the query to the search API provided by the developers of the Synergy task. As expected, the results were poor relative to other submissions.

\begin{table}
    \centering
    \caption{Document retrieval results of the submissions to the BioASQ9 Synergy 2 (top) and BioASQ10 Synergy (bottom) tasks. Metric: F1. The results of rows labelled ``Best'', ``Median'', and ``Worst'' refer to the results of other systems, other than our own, submitted to the challenge.}
    \label{tab:synergydocs}
    \begin{tabular}{lrrrr}
    Run & Round 1 & Round 2 & Round 3 & Round 4\\
    \midrule
    Best&0.3693&0.2039&0.1327&0.1896\\
    Median&0.2388&0.1423&0.0710&0.0800\\
    Worst&0.0157&0.0067&0.0053&0.0175\\
    \MQ-BioASQ9&0.1978&0.1087&0.0483&0.0800\\
    \midrule
    Best&0.3220&0.2221&0.1970&0.1564\\
    Median&0.3100&0.1646&0.1327&0.1067\\
    Worst&0.2729&0.1003&0.0655&0.0478\\
    \MQ-BioASQ10&&0.1003&0.0754&0.0808\\
  \end{tabular}
\end{table}

Table~\ref{tab:synergysnips} shows the F1 score of the snippets returned by our runs. For each run, we indicate the run name, the type of similarity used, and the type of sorting performed. We observe that, considering the poor quality of the documents retrieved, the snippets were of quality comparable to that of other runs of the BioASQ9 Synergy 2 task (but not the runs of the BioASQ10 Synergy task), but there is room for improvement. Among our runs, the most successful configuration was using sBERT cosine similarity and global sort.

\begin{table}
    \centering
    \caption{Snippet retrieval results of the submissions to the BioASQ9 Synergy 2 (top) and BioASQ10 Synergy (bottom) tasks. Metric: F1. The best of our systems in each round is highlighted in \textbf{bold}. The results of rows labelled ``Best'', ``Median'', and ``Worst'' refer to the results of other systems, other than our own, submitted to the challenge.}
    \label{tab:synergysnips}
    \begin{tabular}{lllrrrr}
    Run & Similarity & Sorting & Round 1 & Round 2 & Round 3 & Round 4\\
    \midrule
    Best&&&0.3290&0.1726&0.1262&0.1355\\
    Median&&&0.2288&0.1365&0.0732&0.0764\\
    Worst&&&0.0311&0.0101&0.0231&0.0132\\
    \MQ-1-BioASQ9&tfidf&local&0.1031&0.1035&0.0707&\textbf{0.0764}\\
    \MQ-2-BioASQ9&tfidf&global&0.1100&0.0540&0.0324&0.0619\\
    \MQ-3-BioASQ9&sBERT&local&0.1071&0.0999&0.0692&0.0749\\
    \MQ-4-BioASQ9&sBERT&global&\textbf{0.1923}&\textbf{0.1075}&\textbf{0.1044}&0.0762\\
    \midrule
    Best&&&0.2910&0.1525&0.1574&0.1217\\
    Median&&&0.2757&0.1410&0.1087&0.0948\\
    Worst&&&0.2296&0.0540&0.0273&0.0416\\
    \MQ-1-BioASQ10&tfidf&local&&0.0660&0.0465&0.0771\\
    \MQ-2-BioASQ10&tfidf&global&&0.0540&0.0273&0.0416\\
    \MQ-3-BioASQ10&sBERT&local&&0.0683&0.0457&0.0770\\
    \MQ-4-BioASQ10&sBERT&global&&\textbf{0.0928}&\textbf{0.0725}&\textbf{0.0827}\\
  \end{tabular}
\end{table}

Table~\ref{tab:synergyqa} shows the human evaluation results of the ideal answers returned by our runs. Our runs are very competitive, especially given the relatively poor quality of the input snippets. Given the poor quality of the input snippets in all of our runs, it is dangerous to make generalisations about how the quality of the snippets affect the quality of the answers. Having said that, we can observe that, in the BioASQ9 Synergy 2 task, the runs that generated the best snippets (\MQ-4) did not lead to generating the best ideal answers. The impact of and interplay between the document and snippet retrieval stages, and the question-answering stage, deserves further exploring.

\begin{table}
    \centering
    \caption{Ideal answer results of the submissions to the BioASQ9 Synergy 2 (top) and BioASQ10 Synergy (bottom) tasks. Metric: Average of human evaluation scores. The best of our systems in each round is highlighted in \textbf{bold}. The results of rows labelled ``Best'', ``Median'', and ``Worst'' refer to the results of other systems, other than our own, submitted to the challenge.}
    \label{tab:synergyqa}
    \begin{tabular}{lllrrrr}
    Run & Similarity & Sorting & Round 1 & Round 2 & Round 3 & Round 4\\
    \midrule
    Best&&&4.375&3.850&3.630&3.295\\
    Median&&&3.625&3.100&3.450&3.045\\
    Worst&&&1.042&0.450&3.290&2.060\\
    \MQ-1-BioASQ9&tfidf&local&3.250&3.100&3.450&3.045\\
    \MQ-2-BioASQ9&tfidf&global&3.210&3.075&3.290&\textbf{3.295}\\
    \MQ-3-BioASQ9&sBERT&local&\textbf{3.372}&\textbf{3.250}&\textbf{3.520}&3.067\\
    \MQ-4-BioASQ9&sBERT&global&2.250&3.025&3.490&3.292\\
    \midrule
    Best&&&3.790&3.810&3.562&3.180\\
    Median&&&3.367&3.160&3.250&2.617\\
    Worst&&&3.287&1.550&0.827&0.372\\
    \MQ-1-BioASQ10&tfidf&local&&3.270&3.415&2.617\\
    \MQ-2-BioASQ10&tfidf&global&&3.160&3.305&\textbf{2.990}\\
    \MQ-3-BioASQ10&sBERT&local&&3.360&3.517&2.925\\
    \MQ-4-BioASQ10&sBERT&global&&\textbf{3.490}&\textbf{3.547}&2.690\\
  \end{tabular}
\end{table}

\section{BioASQ10b, Phase B}\label{sec:bioasq10b}

For BioASQ10b, Phase B, we used the question answering system described in Section~\ref{sec:architecture}, using DistilBERT as the BERT variant chosen to compute the word embeddings. Following a data-centric approach, the main difference between the Synergy tasks and BioASQ10, Phase B, is the choice of training data. We hypothesised that the training data that corresponds to the early years of BioASQ, that is, the first samples of the BioASQ10b training data, might be biased. We therefore tested the use of different portions of the training data as shown in Table~\ref{tab:partialtraining}, by incrementally removing the first samples of the training data. We can observe that best evaluation results are obtained with only 50\% of the training data.

\begin{table}
    \caption{Results of 10-fold cross-validation after removing the \emph{first} samples of the BioASQ10b training data. Metric: Average ROUGE-SU4 F1. Best result shown in \textbf{bold}.}
    \label{tab:partialtraining}
    \centering
    \begin{tabular}{l|c}
         Percentage removed & ROUGE-SU4 F1  \\
         \midrule
         10\%&0.281\\ 
         20\%&0.288\\
         30\%&0.298\\
         40\%&0.309\\
         50\%&\textbf{0.311}\\
         60\%&0.308
    \end{tabular}
\end{table}

To double-check that indeed the first samples of the training data are biased, we conducted another round of experiments, but this time removing the last samples of the training data. Table~\ref{tab:partialtraining2} shows that results worsen as the amount of training data diminishes, as one might expect in systems that are based on supervised approaches to machine learning.

\begin{table}
    \caption{Results of 10-fold cross-validation after removing the \emph{last} samples of the BioASQ10b training data. Metric: Average ROUGE-SU4 F1.}
    \label{tab:partialtraining2}
    \centering
    \begin{tabular}{l|c}
         Percentage removed & ROUGE-SU4 F1  \\
         \midrule
         10\%&0.275\\ 
         20\%&0.268\\
         30\%&0.270\\
         40\%&0.255\\
         50\%&0.241\\
         60\%&0.229
    \end{tabular}
\end{table}

Hyperparameter search showed that the same hyperparameters give optimal results when training using the entire training data, or using only 50\% of the training data: dropout=0.6, number of epochs=1.

\subsection{Submission Results to BioASQ10b, Phase B}

Table~\ref{tab:bioasq10b} shows the results of our submissions to BioASQ10b, Phase~B\footnote{At the time of writing, only the automated evaluation results were available.}. Note that the results reported in the BioASQ website\footnote{\url{http://bioasq.org}} may change in the future after the test data is potentially enriched with further annotations.

\begin{table}
    \centering
    \caption{Preliminary results of the submissions to BioASQ10b, Phase B. The best of our systems in each batch is highlighted in \textbf{bold}. The results of rows labelled ``Best'', ``Median'', and ``Worst'' refer to the results of all systems, including our own, submitted to the challenge.}
    \label{tab:bioasq10b}
   \begin{tabular}{llrrrrrr}
     && \multicolumn{5}{c}{ROUGE-SU4 F1}\\
     Run&Training Data & Batch 1 & Batch 2 & Batch 3 & Batch 4 & Batch 5 & Batch 6\\
     \midrule
     Best&&0.3715&0.4168&0.3689&0.4165&0.3916&0.1705\\
     Median&&0.3339&0.3521&0.3387&0.3556&0.3389&0.1581\\
     % Worst&&\\
     \midrule
     \MQ-1&All BioASQ10b&\bf 0.3490&\bf 0.3484&\bf 0.3344&0.3525&0.3415&0.1581\\
     \MQ-2&Last 50\% of BioASQ10b&0.3339&0.3480&0.3316&\bf 0.3556&\bf 0.3431&\bf 0.1640\\
   \end{tabular}
\end{table}

Our runs are comparable to the median of those of other participating systems. Surprisingly, there is little difference between using all training data or only the latter 50\%. When we visually inspected the outputs of the runs, we noticed that the output of all runs in each batch were virtually identical, with only a few differences.

\section{Summary and Conclusions}\label{sec:conclusions}

We have presented \MacquarieUniversity's contribution to the BioASQ9 Synergy task 2, the BioASQ10 Synergy task, and BioASQ10b, Phase~B (Ideal Answers). In all of our runs, the base question answering architecture was virtually the same, the only differences being the choice of DistilBERT vs. ALBERT, and the training data used.

For the synergy tasks, we used a system that has been trained using BioASQ9b training data. We experimented with approaches for snippet retrieval based on two dimensions: vectors used for similarity comparison, and final ranking approach. Cosine similarity using sBERT gave the best results, and we observed that not always the best snippets for the snippet retrieval task led to best answers in the question answering task.

Overall, the results of the question answering parts were competitive, especially given the relatively poor quality of the documents and snippets retrieved. We will investigate approaches to increase the quality of the retrieval stages, and explore the relation between quality of retrieval vs. quality of final answers.

For the BioASQ10b, Phase~B task, we followed a data-centric approach and experimented with training regimes that incrementally removed samples from the training data. During our preliminary cross-validation experiments we observed an improvement of results using only the latter 50\% of the training data, but this difference of results vanished in the submitted runs.

With a data-centric approach in mind, we plan to conduct further experiments that test the impact of changes and transformations of the training data. For example, besides further examining the impact of using portions of the training data, we will investigate the use of data augmentation techniques.

\begin{acknowledgments}

This research was undertaken with the assistance of resources and services from the National Computational Infrastructure (NCI), which is supported by the Australian Government.

\end{acknowledgments}

%%
%% Define the bibliography file to be used
\bibliography{bioasq}

%\appendix

\end{document}